# Machine Learning based Prediction of Hierarchical Classification of Transposable Elements

Manisha Panta*
Department of Computer Science
The University of New Orleans
New Orleans, LA, United States
mpanta1@uno.edu

Avdesh Mishra*
Department of Computer Science
The University of New Orleans
New Orleans, LA, United States
amishra2@uno.edu

Md Tamjidul Hoque†
Department of Computer Science
The University of New Orleans
New Orleans, LA, United States
thoque@uno.edu

Joel Atallah†
Department of Biological Sciences
The University of New Orleans
New Orleans, LA, United States
jatallah@uno.edu

## ABSTRACT

Transposable Elements (TEs) or jumping genes are the DNA sequences that have an intrinsic capability to move within a host genome from one genomic location to another. Studies show that the presence of a TE within or adjacent to a functional gene may alter its expression. TEs can also cause an increase in the rate of mutation and can even mediate duplications and large insertions and deletions in the genome, promoting gross genetic rearrangements. Thus, the proper classification of the identified jumping genes is essential to understand their genetic and evolutionary effects in the genome. While computational methods have been developed that perform either binary classification or multi-label classification of TEs, few studies have focused on their hierarchical classification. The state-of-the-art machine learning classification method utilizes a Multi-Layer Perceptron (MLP), a class of neural network, for hierarchical classification of TEs. However, the existing methods have limited accuracy in classifying TEs. A more effective classifier, which can explain the role of TEs in germline and somatic evolution, is needed. In this study, we examine the performance of a variety of machine learning (ML) methods. And eventually, propose a robust approach for the hierarchical classification of TEs, with higher accuracy, using Support Vector Machines (SVM).

*These authors contributed equally to this work as first authors.
†To whom Correspondence should be addressed.

## 1 INTRODUCTION

Transposable Elements (TEs) are repetitive genomic sequences. TEs are DNA sequences that have the intrinsic capability to move within a host genome from one genomic location to another. Barbara McClintock identified transposons or jumping genes [1], and she introduced the concept of transposons through an analysis of genetic instability in the inheritance of pigmentation in maize [1]. TEs make up a substantial fraction of the host genome in which they reside. Genome sequencing projects have shown that TEs make 25%-50% of mammalian genomes [2]. Numerous recent studies on the identification and classification of TEs, along with their effects in the genome, show that TE's are not just "Junk DNA". On the contrary, they are responsible for genetic variability, modifying gene function and expression, in addition to increasing the size of the genome [2-5]. As they move from one position to another in the genome, they cause an assortment of genetic instabilities, including mutations and chromosome breakage [5]. Therefore, the proper classification of TEs is vital to understand their specific role in germline and somatic evolution. To perform the classification of transposable elements, several tools are available [6-12]. However, the accuracy of the available tools is often low . Therefore, we proposed to build an effective machine learning model that can hierarchically classify transposable elements up to the superfamily level of the Wicker's taxonomy [13], potentially opening the door to the identification of the role of TEs in genome evolution with higher confidence.

Historically, the classification of TEs has been based mainly on their mechanism of transposition ("copy-and-paste" vs. "cut-and-paste") together with the comparison of their genomic structures and sequence similarities. On this basis, a taxonomy known as the "unified classification system for eukaryotic transposable elements", proposed by Wicker et al. [13], exploits the hierarchical relationships between classes of TEs. This taxonomy has been extensively utilized in the development of several automated TE classification tools [6-9]. The detailed classification used in this research is presented in Figure 1.

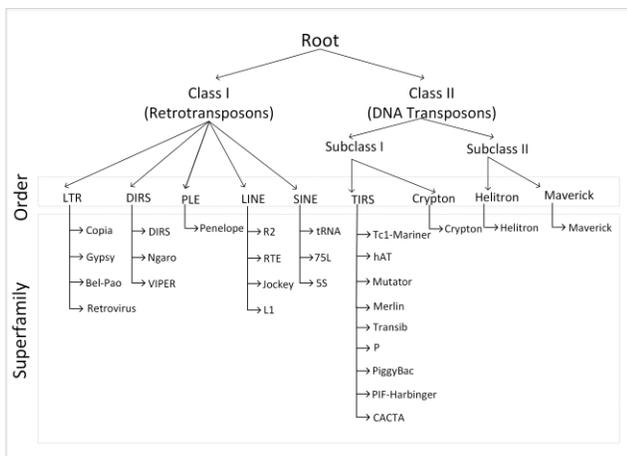

**Figure 1: Taxonomy of TEs proposed by Wicker et.al [13]**

The rest of this paper is organized as follows. In section 2, we review some relevant tools and research works related to the hierarchical classification of the TEs. In section 3, we describe the experimental setups of our study. Here, we introduce the datasets, feature extraction procedure, hierarchical classification techniques, evaluation metrics used in our work and the motivation behind choosing a hierarchical classification approach. Section 4 includes elaboration on the parameter selection and optimization of several state-of-the-art machine learning (ML) techniques implemented and compared in this study. Section 5 presents the results of different ML techniques, including the performance comparison to relevant state-of-the-art technique. Finally, section 6 concludes the research work with the selection of the best performing predictor framework and future directions.

## 2 BACKGROUND

As TEs are present in abundance in the genome, systematic organization of them into a hierarchical structure is important so that the taxonomy can be easily applied by both experts and non-experts in research studies [13]. Hierarchical classification (HC) of TEs exploits hierarchical relationships between classes, simplifying the understanding of their intrinsic and extrinsic characteristics which, successively help to identify unknown traits of new sequence based on the classified sequences.

HC consists of classification problems whose classes are organized in a predefined hierarchy or taxonomy [14]. The taxonomy can either be represented by a directed graph (DAG) or a treelike structure with class labels as nodes. In [14], the authors formally defined the hierarchical classification problem as a 3-tuple (Y, Ψ, Φ), where, Y specifies the hierarchy (either treelike or DAG), Ψ specifies whether the instance has single path of label prediction or multiple paths of label prediction and finally, Φ specifies whether an instance has full depth labeling (mandatory leaf node prediction) or partial (non-mandatory leaf node prediction).

Various approaches and tools have been developed in the process of solving hierarchical classification problems using machine learning. Real-world problems addressed using HC methods include text classification [15], protein function prediction [16] and classification of web content [17]. In the state-of-the-art study on TE classification [12], the authors introduced a hierarchical classification method to classify TEs using Multi-Layer Perceptron (MLP), a class of neural networks as a machine learning technique. Using the publicly available repositories, the authors assembled hierarchical TE datasets that can be used to validate machine learning models. Additionally, they proposed two hierarchical classification strategies: non-Leaf Local Classifier per Parent Node (nLLCPN) and Local Classifier per Parent Node and Branch (LCPNB) for the specific purpose of a hierarchical classification of TEs. As reported, these strategies were shown to be statistically competitive or even superior to earlier hierarchical strategies. Both nLLCPN and LCPNB allow one local binary or multi-class classifier per node of the class hierarchy (except the root node) to make non-mandatory leaf-node predictions by replicating the internal node as a subclass of itself. However, the LCPNB approach tries to avoid error propagation by taking advantage of local information, such as the use of prediction probabilities to predict the final class. The same authors, in a separate study [18] implemented Denoising Auto-Encoder (DAEs) and Deep MLP (DMLP) to improve the performance of HC methods and presented a level-wise assessment of model performance.

## 3 EXPERIMENTAL SETUP

In this section, we describe our approach for training and validation-dataset collection, feature extraction, hierarchical classification strategies, performance evaluation metrics and finally, the approach we took to establish the machine learning framework for the hierarchical classification of TEs.

### 3.1 Dataset Collection

Three hierarchical datasets previously established by Nakano *et. al.* [12], are used in this study to train and validate the machine learning framework. Two out of three hierarchical datasets were extracted from Plant Genome and System Biology (PGSB) [19] and, REPBASE [4], respectively, and the third dataset (PGSB + REPBASE) was created by combining the PGSB and REPBASE. The repetitive DNA sequences from PGSB and REPBASE public repositories were first preprocessed, and then, hierarchically organized as a tree and labeled according to the taxonomy proposed by Wicker *et. al.* [13]. Each sequence collected from both the repositories is numerically labeled. The numerical label of the TE class represents its position in the hierarchy in the Wicker's taxonomy. For example, if a TE sequence is labeled as *Copia*, it is numerically labeled as (1.1.1), which represents that Copia (1) is a superfamily of order LTR (1), which in turn is a subclass of Class I (1) retrotransposons.

Table 1 shows some of the properties of the datasets that we used in our study. The first column of the table gives the total number of instances in the individual dataset, the second column is about the total number of features used to train our machine learning models, and the third column provides information about the number of classes available per level in each dataset.

**Table 1: Dataset Statistics.**

| Datasets | Number of Instances | Number of Features | Classes Per Level |
|---|---|---|---|
| PGSB | 18678 | 336 | 2 / 4 / 3 / 5 |
| REPBASE | 34559 | 336 | 2 / 5 / 12 / 9 |
| PGSB + REPBASE | 53049 | 336 | 2 / 5 / 12 / 9 |

### 3.2 Feature Extraction

Feature extraction is a significant step in determining a robust model. The k-mer frequency count in bioinformatics is simple in principle yet can reveal necessary information about a fasta sequence, especially about the relative distribution of substrings within the nucleotide sequence. An example of the feature extraction procedure is shown in Figure 2.

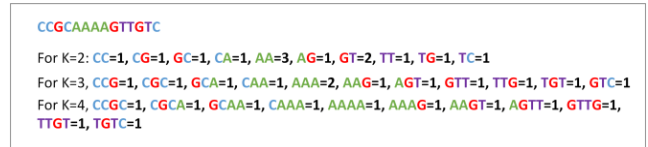

**Figure 2: An example of the feature extraction procedure from a sequence.**

In this study, k-mers with $K=2$, $K=3$, and $K=4$ were used to generate a total of 336 features [12], which have been used in training the model. Counting k-mers of each sequence was performed using a computational tool known as BFCounter [20]. BFCounter uses a bloom filter, which is a probabilistic data structure to store all the observed k-mers. A total of 336 features (k-mers) were generated using this method of feature extraction. Table 2 shows the statistics of the features used in the study.

**Table 2: Types and Number of Features**

| Feature Type | Number of Features |
|---|---|
| $K = 2$ | 16 |
| $K = 3$ | 64 |
| $K = 4$ | 256 |

### 3.3 Hierarchical Classification Strategies

Here, we discuss two state-of-the-art top-down hierarchical classification strategies that we used in this study to classify TEs hierarchically. Non-Leaf Local Classifier per Parent Node (nLLCPN) and Local Classifier per Parent Node and Branch (LCPNB) the two hierarchical classification approaches proposed in the paper [12]. These algorithms were specifically designed and tested to avoid error propagation during training and testing of models for hierarchical classification of TEs. Because of the nature of the TE datasets being generated with labels that do not support mandatory-leaf node prediction, the algorithms were designed to allow predictions for non-mandatory leaf node prediction, i.e., the prediction from the intermediate node is also treated as a valid prediction. A classifier in each parent node learns to distinguish among its sub-classes and itself by allowing a parent node to add an extra node to itself as a child node, consequently, supporting non-mandatory leaf node prediction.

### 3.3.1 non-Leaf Local Classifier per Parent Node (nLLCPN)

In nLLCPN, a multiclass classifier is trained for each parent node or non-leaf node of the hierarchy. During the testing phase, all the trained local classifiers were stacked as a hierarchy of a flat classifier and using a top-down prediction approach was made. It follows the top-down classification approach in which the path with higher probabilities gives the final route of classification. The non-mandatory leaf node prediction is the stopping criteria for this classification approach. Under this criterion, the classification of an instance can be stopped at the internal node as a classifier associated with an internal node predicts itself. So, to achieve internal leaf node prediction, the algorithm nLLCPN replicates itself as a child node of its own.

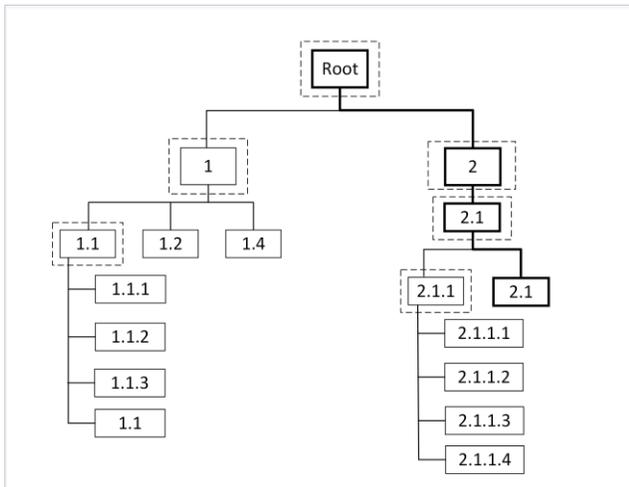

Figure 3: Classification using nLLCPN Hierarchical Classification strategy.

Figure 3 illustrates how nLLCPN supports non-mandatory leaf node prediction. A trained classifier is associated with every parent node (dashed rectangular boxes in the figure), and the bold line represents the path of final classification of a sample TE. Note that if an instance may be classified, for example, as class 2.1, but does not belong to class 2.1.1 then, nLLCPN replicates itself as a child node of its own such that the instance can be classified as 2.1 as can be seen in figure 3.

### 3.3.2 Local Classifier per Parent Node and Branch (LCPNB)

In the LCPNB classification strategy, the training phase is the same as that used in the nLLCPN strategy, i.e., a multiclass classifier was trained at each parent node or non-leaf node of the hierarchy. LCPNB supports non-mandatory leaf node prediction in a different way. During the classification or testing phase, a new instance is provided as an input to the trained classifier of every parent node, and the prediction probabilities were acquired for all the classes. Then the average probabilities of all the possible paths from the root to the nodes representing classes were calculated. The final classification of an instance was then chosen as the path with the highest average probability value.

Figure 4 illustrates the classification using the LCPNB strategy. A trained classifier associated with every parent node is shown by the dashed rectangular box in the figure, the value inside each bracket is the prediction probability associated with the particular class, and the bold line represents the path of final classification of a sample TE.

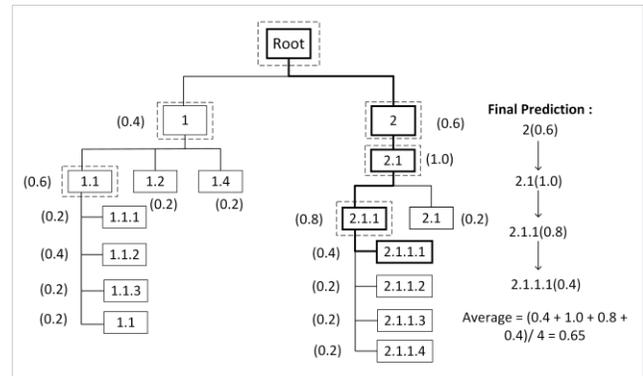

Figure 4: Classification using LCPNB Hierarchical Classification strategy.

### 3.4 Performance Evaluation

In this section, we discuss the different ways in which we evaluated the performance of the different machine learning models. The performance evaluation of the hierarchical classification of TEs can be measured by using standard metrics but modified to evaluate the classification at the hierarchical levels. In this research, we used hierarchical performance metrics, which were highly recommended in the paper by Silla and Freitas [14]. These hierarchical metrics are hierarchical Precision (hP), hierarchical Recall (hR) and hierarchical F-measure (hF) and are defined by formulas 1, 2, and 3, respectively. The datasets obtained from the state-of-the-art method had been systematically split into 10 train and 10 test datasets using a stratified 10-fold CV. We implemented the loop to train our classifiers with all those 10 training datasets. Then tested each trained model with their respective test dataset to compute the values of performance metrics using formulas 1, 2, and 3. We then recorded the average over 10 executions.

$$hP = \frac{\sum_i |P_i \cap T_i|}{\sum_i |P_i|} \quad (1) \quad hR = \frac{\sum_i |P_i \cap T_i|}{\sum_i |T_i|} \quad (2) \quad hF = \frac{2 * hP * hR}{hp + hR} \quad (3)$$

Here, $P_i$ is a set of predicted class(es) and $T_i$ is a set of true class(es) for a test sample $i$.

## 4 MACHINE LEARNING APPROACHES

In this section, we elaborate on the parameter selection and optimization for all the individual state-of-the-art machine learning algorithms.

### 4.1 Learning Algorithms

According to the *No Free Lunch* theorem developed by Wolpert *et. al.* [21], in the machine learning aspect, it can be surmised that every problem is unique, and no specific algorithm works best for every problem. The selection of learning algorithm depends on the size and the quality of data, roughness of the decision boundary, computational time to be considered, and the problem definition. Henceforth, we tried different algorithms for our hierarchical classification problem and then evaluated the performance of each model using hierarchical evaluation metrics. We then proposed a novel SVM-based ML framework for hierarchical classification of transposable elements.

We investigated eight supervised machine learning algorithms in different parameter settings. We used simple yet sometime much practical algorithms such as k-Nearest Neighbors (kNN) [22], Logistic Regression (LogReg) [23], ensemble algorithms based on bagging and boosting such as Random Forest (RF) [24], Extremely Randomized Trees (ET) [25], Gradient Boosting Classifier (GBC) [26], eXtreme Gradient Boosting Classifier (XGBC) [27], Multi-Layer Perceptron (MLP) [28] and Support Vector Machines (SVM) [29].

We used the Scikit-learn library [30] to build models and tune the parameters of all of the aforementioned learning algorithms. We used the Hit-and-Trial method for parameter selection for some machine learning algorithms, whereas for some ML algorithms, we used the default parameter settings. Since not all algorithms provide significant improvement with hyperparameter optimization, we implemented optimization only for SVM. SVM with the appropriate kernel can have better accuracy with the prediction, especially if the dataset is non-linearly separable and the problem domain comprises of high-dimensional space. SVM being powerful and popular in pattern recognition and classification problems in many fields such as bioinformatics [31], image classification [32], and text classification [33], perhaps provided us confidence in focusing it on being a predictor framework.

As parameters, for KNN, we used $k = 15$, for ET and RF, 1000 estimators with the maximum depth of 8 were used and for GBC 2000 estimators, 0.2 learning rate, and max_depth of 8 were used. For LogReg, and XGBC default parameter settings were used. Likewise, MLP with one hidden layer with 200 nodes with logistic activation function was implemented the same as that used in the state-of-the-art method. Each learning algorithm was trained on all the three datasets and tested using two classification strategies i.e., LCPNB and nLLCPN, with a 10-fold cross-validation strategy. The prediction abilities of each model were then evaluated using hierarchical performance metrics.

### 4.2 Predictor Framework

Support Vector Machine (SVM) [29] with radial basis function (RBF) kernel is used to design a predictor framework to perform a multi-class classification of transposable elements at each level of the hierarchy. We identified the proper combination of cost (C) and gamma ($\gamma$) parameters of SVM with RBF kernel to achieve a better classification performance of the predictor. To optimize the RBF kernel parameter ($\gamma$) and the cost parameter C, we used a grid search technique [34]. We identified the optimal values of parameters for each of the three datasets by grid search using a 10-fold cross-validation technique.

In the training phase, we generated the subsets of the dataset for all parent nodes (note that each subset of a dataset contains feature-set with all its child nodes) and invoked our SVM with optimized parameters. So, for each fold of a training dataset, every parent node in the graph, an SVM model was developed.

In the testing phase, for each fold of the test dataset, we used both nLLCPN and LCPNB hierarchical classification strategies to test our SVM based predictor. The model was evaluated based on the predictions using hierarchical evaluation metrics, and the average of the metrics for 10-fold test datasets was recorded.

# 5 RESULTS

In the first part of this section, we present a comparative analysis of the two hierarchical classification methods that we used in the study tested using different machine learning methods. And in the next part of this section, we present the performance of our proposed SVM-based method, including its comparison to a class of neural network, MLP, the learning algorithm used in the state-of-the-art method.

## 5.1 Performance Comparisons of Classification Strategies

The results in this section are from the experiments we performed on two hierarchical classification strategies for transposable elements - non-Leaf Local Classifier per Parent Node (nLLCPN) and Local Classifier per Parent Node and Branch (LCPNB). We analyzed the performances of LCPNB and nLLCPN using eight different machine learning models. The models were trained on three different datasets PGSB, REPBASE, and Combined of PGSB and REPBASE. We found that performance of LCPNB strategy is competitive to or higher than that of nLLCPN.

Figures 5, 6, and 7 show the comparison between the prediction outputs of the learners for two classification strategies. LCPNB provided better or same hierarchical F-measure for all the learners, especially for Multi-layer Perceptron (a type of neural network), Randomized Forest (RF), and Support Vector Machine (SVM).

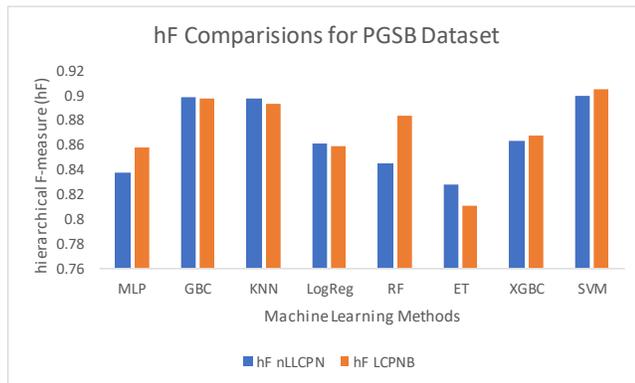

**Figure 5: Shows comparative results of different machine learning approaches in the PGSB hierarchical dataset.**

As recommended in the paper [12] and from this result, we can, with higher confidence, state that the LCPNB strategy has an overall significant impact on the hierarchical classification of transposable elements. The superior performance of the algorithm is obtained because the average probabilities of all the paths are computed before making a final classification of an instance.

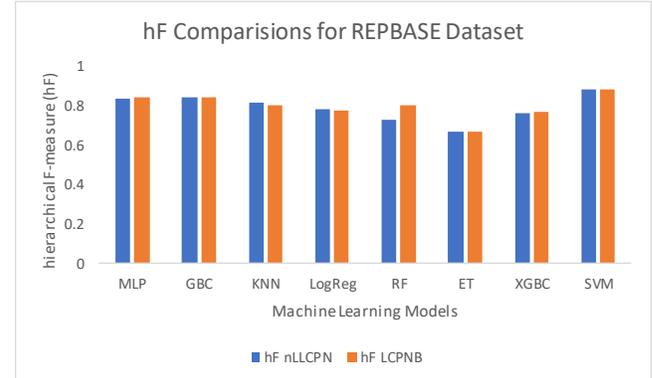

**Figure 6: Shows comparative results of different machine learning approaches in the REPBASE hierarchical dataset.**

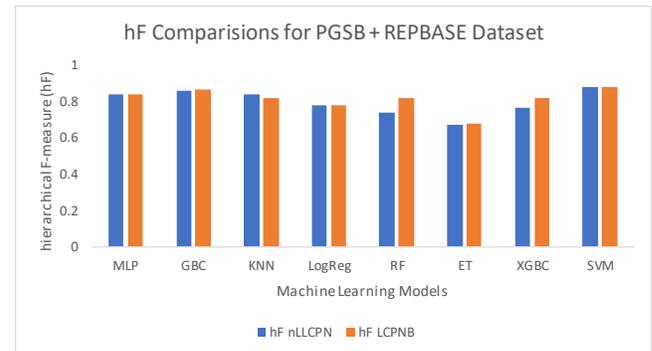

**Figure 7: Shows comparative results of different machine learning approaches in the Mixed hierarchical dataset.**

## 5.2 Performance Evaluation of Predictor Model

In this section, we evaluate the performance of the optimized SVM based model for three datasets and compare it with the state-of-the-art method. Table 3 shows the overall performance of the predictor framework for two hierarchical classification strategies on three datasets. The values of hierarchical Recall (hR), hierarchical Precision (hP) and hierarchical F-measure (hF) for three datasets on two hierarchical classification strategies can be observed in Table 3. Likewise, Table 4 shows the results of the SVM model based on level-wise prediction performance. The optimized SVM with RBF kernel gives an outstanding hierarchical F-measure (hF) as compared to the state-of-the-art method. From Table 5, we observe our predictor presented percentage improvement with respect to the MLP-based state-of-the-art method.

**Table 3: Performance of SVM-based predictor.**

| METRICS | PGSB | REPBASE | PGSB + REPBASE |
|---|---|---|---|
| | nLLCPN | | |
| hR | 0.897 | 0.887 | 0.879 |
| hP | 0.908 | 0.879 | 0.882 |
| hF | 0.903 | 0.883 | 0.881 |
| | LCPNB | | |
| hR | 0.904 | 0.890 | 0.884 |
| hP | 0.907 | 0.881 | 0.880 |
| hF | 0.905 | 0.885 | 0.882 |

**Table 4: Level-wise results of the hierarchy for three datasets.**

| | | hF nLLCPN | hF LCPNB |
|---|---|---|---|
| PGSB | Level 1 | 0.949 | 0.950 |
| | Level 2 | 0.943 | 0.945 |
| | Level 3 | 0.852 | 0.857 |
| | Level 4 | 0.569 | 0.618 |
| REPBASE | Level 1 | 0.958 | 0.959 |
| | Level 2 | 0.949 | 0.950 |
| | Level 3 | 0.869 | 0.872 |
| | Level 4 | 0.746 | 0.753 |
| PGSB + REPBASE | Level 1 | 0.952 | 0.952 |
| | Level 2 | 0.943 | 0.943 |
| | Level 3 | 0.852 | 0.854 |
| | Level 4 | 0.696 | 0.705 |

Theoretically we know that SVM algorithm implemented using an appropriate kernel trick works well in high dimensional feature spaces. Also, optimizing hyperparameters of a machine learning method provides better performance of the predictor. Since we chose best performing kernel trick and tuned the hyperparameters using computationally exhaustive grid-search, we were able to generate the best performing predictor model using SVM with RBF kernel.

**Table 5: Percentage improvement of proposed method.**

| | PGSB | REPBASE | PGSB+ REPBASE |
|---|---|---|---|
| | nLLCPN | | |
| MLP | 0.848 | 0.838 | 0.839 |
| Proposed SVM | 0.903 | 0.883 | 0.881 |
| imp. % | 7.40% | 5.40% | 5% |
| | LCPNB | | |
| MLP | 0.858 | 0.848 | 0.845 |
| Proposed SVM | 0.905 | 0.885 | 0.882 |
| imp. % | 5.48% | 4.48% | 4.26% |

**Here, 'imp. %' indicates percentage improvement achieved by our proposed method over the respective state-of-the-art method.**

## 6 CONCLUSIONS

In this paper, we first compared two hierarchical classification approaches, nLLCPN, and LCPNB and found out that the performance of the LCPNB approach is competitive or superior with most of the machine learning methods. So, we recommend LCPNB hierarchical classification strategy to be used in future experiments. Secondly, we implemented eight machine learning approaches with the aim of determining individual ML algorithms that can provide the best prediction of the hierarchical classes of a transposable element. The balanced accuracy of our proposed optimized SVM with RBF kernel outperformed all the machine learning algorithms. Consequently, the proposed SVM based method can recognize the class of transposable elements in a hierarchy with better confidence. The tool based on such higher confidence is helpful for the researchers in identifying and analyzing the role of TEs in genome evolution.